\documentstyle[twoside,palatino,cl]{article}

\input{psfig}

\newcounter{algcounter}
\renewcommand{\thealgcounter}{\arabic{algcounter}}
\newcommand{\algcounter}{\refstepcounter{algcounter}\thealgcounter}

\newcommand{\functionset}[2]{{#1} \longrightarrow {#2}}
\newcommand{\signaturecolon}[3]{{#1} : \functionset{#2}{#3}}
\newcommand{\funclang}{{\mathcal{L}}}
\newcommand{\quanttwo}[2]{\{\,#1 \mid #2\,\}}
\newcommand{\funclangright}{\stackrel{\rightarrow}{\cal L}}
\newcommand{\powerset}[1]{{\cal P}(#1)}
\newcommand{\rpred}[1]{\mathit{#1}}
\newcommand{\quantforall}[2]{\forall_{#1}#2}
\newcommand{\quantexists}[2]{\exists_{#1}#2}
\newcommand{\rrm}[1]{{\rm #1}}
\newcommand{\nametrue}{{\em true}}
\newcommand{\emptystring}{\varepsilon}
\newcommand{\implies}{\Rightarrow}
\newcommand{\rmathfunc}[1]{\mbox{{\it #1}}}
\newcommand{\bigoh}[1]{{\cal O}(#1)}
\newcommand{\rvar}[1]{{\it #1}}

\title{Incremental Construction of Minimal Acyclic Finite-State Automata}

\author{
    Jan Daciuk
    & Stoyan Mihov \\
    \affil{Technical University of Gda\'nsk}%
        \thanks{Department of Applied Informatics,
          Technical University of Gda\'nsk,
        Ul.\ G.\ Narutowicza 11/12,
        PL80-952 Gda\'nsk, Poland; e-mail: jandac@pg.gda.pl
        }
    & \affil{Bulgarian Academy of Sciences}%
        \thanks{Linguistic Modelling Laboratory,
        LPDP --- Bulgarian Academy of Sciences, Bulgaria;
        e-mail: stoyan@lml.bas.bg
        } \\
      \ \\
    Bruce W.\ Watson
    & Richard E.\ Watson%
        \thanks{e-mail: watson@OpenFIRE.org} \\
    \affil{University of Pretoria}%
        \thanks{Department of Computer Science, University of Pretoria,
        Pretoria 0002, South Africa, e-mail:
        watson@cs.up.ac.za
        }
    &
  }

\runningtitle{\hfill{}Incremental Construction of FSAs}
\runningauthor{Daciuk and Mihov and Watson and Watson}
\issue{26}{1}{2000}
\date{}
\begin{document}
\maketitle

\begin{abstract}
  In this paper, we describe a new method for constructing minimal,
  deterministic, acyclic finite-state automata from a set of strings.
  Traditional methods consist of two phases: the first to construct
  a trie, the second one to minimize it. Our approach is to
  construct a minimal automaton in a single phase by adding new
  strings one by one and minimizing the resulting automaton
  on-the-fly. We present a general algorithm as well as a
  specialization that relies upon the lexicographical ordering of the
  input strings.
  Our method is fast and significantly lowers memory requirements
  in comparison to other methods.
\end{abstract}

\section{Introduction}
Finite state automata are used in a variety of applications, including aspects
of
natural language processing (NLP). They may store sets of words,
with or without annotations such as the corresponding
pronunciation, base form, or morphological categories. The main reasons
for using finite state automata in the NLP domain are that their
representation of the set of words is compact and that looking up a string in
a dictionary represented by a finite-state automaton is very fast  ---
proportional to the length of the string. Of particular interest to
the NLP community are deterministic,
acyclic, finite-state automata, which we call {\bf dictionaries}.

Dictionaries can be constructed in various ways ---
see Watson~\shortcite{Wats93a,Wats95} for a taxonomy of (general) finite
state automata construction algorithms. A word is simply a finite sequence of
symbols over some alphabet and we do not associate it with a meaning in this
paper.
A necessary and sufficient condition for any deterministic automaton
to be acyclic is that it recognizes a finite set of words.  The
algorithms described here construct automata from such finite sets.

The Myhill-Nerode theorem (see \namecite{HU79}) states that
among the many deterministic
automata that accept a given language, there is a unique automaton
(excluding isomorphisms) that has a minimal number of states. This is
called the {\bf minimal} deterministic automaton of the language.

The generalized algorithm presented in this paper has been independently
developed by Jan Daciuk of the Technical University of
Gda\'nsk, and by Richard Watson and Bruce Watson (then of the IST
Technologies Research Group) at Ribbit Software Systems Inc.
The specialized (to
sorted input data) algorithm was independently developed by Jan Daciuk and by
Stoyan Mihov of the Bulgarian Academy of Sciences. Jan Daciuk
has made his C++ implementations of the algorithms freely available for
research purposes at
www.pg.gda.pl/$\sim$jandac/fsa.html.\footnote{The algorithms in
Daciuk's implementation differ slightly from those presented here, as he uses
automata with {\em final transitions}, not final states. Such automata have
fewer states and fewer transitions than traditional ones.}
Stoyan Mihov has implemented the (sorted input) algorithm in a Java
package for minimal acyclic finite-state automata. This package forms
the foundation of the Grammatical Web Server for Bulgarian (at
origin2000.bas.bg) and implements operations on acyclic finite
automata, such as union, intersection and difference, as well as
constructions for perfect hashing.
Commercial C++ and Java
implementations are available via www.OpenFIRE.org. The commercial
implementations include several additional features such as a method
to remove words from the dictionary (while maintaining minimality).
The algorithms have been used for constructing
dictionaries and transducers for spell checking, morphological
analysis, two-level morphology, restoration of diacritics, perfect
hashing, and document indexing.  The algorithms have also
proven useful in numerous problems outside the field of NLP, such as
DNA sequence matching and computer virus recognition.

An earlier version of this paper, authored by Daciuk, Watson, and Watson,
appeared at the International Workshop on Finite-state Methods in Natural
Language Processing in 1998 --- see \namecite{DWW98}.

\section{Mathematical Preliminaries}
We define a deterministic finite state automaton to be a
5-tuple $M=(Q, \Sigma, \delta, q_0, F)$, where $Q$ is a finite set of
states, $q_0 \in Q$ is the start state, $F \subseteq Q$ is a set of
final states, $\Sigma$ is a finite set of symbols called the {\em alphabet\/}
and $\delta$ is a partial mapping $\signaturecolon{\delta}{Q \times \Sigma}{Q}$
denoting transitions. When $\delta(q,a)$ is undefined, we write $\delta(q,a) =
\bot$.
We can extend the $\delta$ mapping to partial mapping
$\signaturecolon{\delta^*}{Q \times \Sigma^*}{Q}$ as follows (where
$a \in \Sigma$, $x \in \Sigma^*$):
\begin{eqnarray*}
    \delta^*(q,\emptystring)    &   =   &   q \\
    \delta^*(q,ax)              &   =   &   \left\{ \begin{array}{ll}
                                                    \delta^*(\delta(q,a),x) &   \mbox{if $\delta(q,a) \neq \bot$} \\
                                                    \bot                    &   \mbox{otherwise}
                                                    \end{array}
                                            \right.
\end{eqnarray*}
Let DAFSA be the set of all deterministic finite state automata in which
the transition function $\delta$ is acyclic --- there is no string $w$
and state $q$ such that $\delta^*(q,w) = q$.

We define $\funclang(M)$ to be the
language accepted by automaton $M$:
\[
    \funclang(M) = \quanttwo{x \in \Sigma^*}{\delta^*(q_0,x) \in F}
\]
The size of the automaton, $|M|$, is equal to the number of states,
$|Q|$. $\powerset{\Sigma^*}$ is the set of all languages over $\Sigma$.
Define the function $\signaturecolon{\funclangright}{Q}{\powerset{\Sigma^*}}$
to map a state $q$ to the set of all strings on a path from $q$ to any final
state in $M$. More precisely,
\[
    \funclangright(q) = \quanttwo{x \in \Sigma^*}{\delta^*(q,x) \in F}
\]
$\funclangright(q)$ is called the {\bf right language} of $q$.
Note that $\funclang(M) = \funclangright(q_0)$.
The right language of a state can also be defined recursively:
\[
    \funclangright(q) = \quanttwo{a \funclangright(\delta(q,a))}{a \in \Sigma \land \delta(q,a) \neq \bot}
                            \cup  \left\{  \begin{array}{cl}
                                            \{\emptystring\}    &   \mbox{if $q\in F$} \\
                                            \emptyset           &   \mbox{otherwise}
                                            \end{array}
                                  \right.
\]
One may ask whether such a recursive definition has a unique solution. Most
texts on language theory, for example Arbib, Moll and Kfoury \shortcite{mak},
show that the solution is indeed unique --- it is the least fixed-point of the
equation.

We also define a property of an automaton specifying that all states can be
reached from the start state:
\[
    \rpred{Reachable}(M) \equiv
        \quantforall{q \in Q}{
            \quantexists{x \in \Sigma^*}{(\delta^*(q_0, x) = q)}
        }
\]
The property of being a minimal automaton is traditionally defined as
follows (see Watson~\shortcite{Wats93b,Wats95}):
\[
    \rpred{Min}(M) \equiv
        \quantforall{M' \in \rrm{DAFSA}}{(\funclang(M) = \funclang(M')
            \implies |M| \leq |M'|)}
\]
We will, however, use an alternative definition of minimality, which is shown
to be equivalent:
\[
\rpred{Minimal}(M) \equiv (\quantforall{q, q' \in Q}{ (q \neq q'
    \implies
    \funclangright(q) \neq \funclangright(q'))}) \land \rpred{Reachable}(M)
\]
A general treatment of automata minimization can be found in
Watson~\shortcite{Wats95}.
A formal proof of the correctness of the following
algorithm can be found in~\namecite{mihov98}.

\section{Construction from Sorted Data}
\begin{flushleft}
\begin{figure}[htb]
\mbox{\psfig{figure=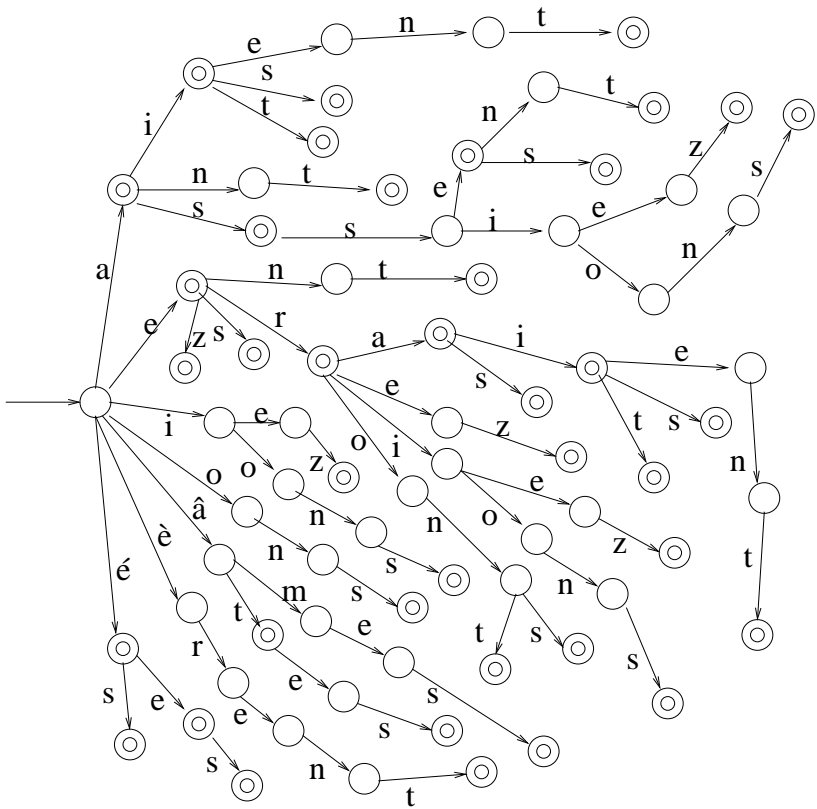,width=0.8\textwidth}}
\caption{A trie whose language is the French regular endings of verbs
  of the first group.}%
\label{fig:before}
\end{figure}
\begin{figure}[htb]
\mbox{\psfig{figure=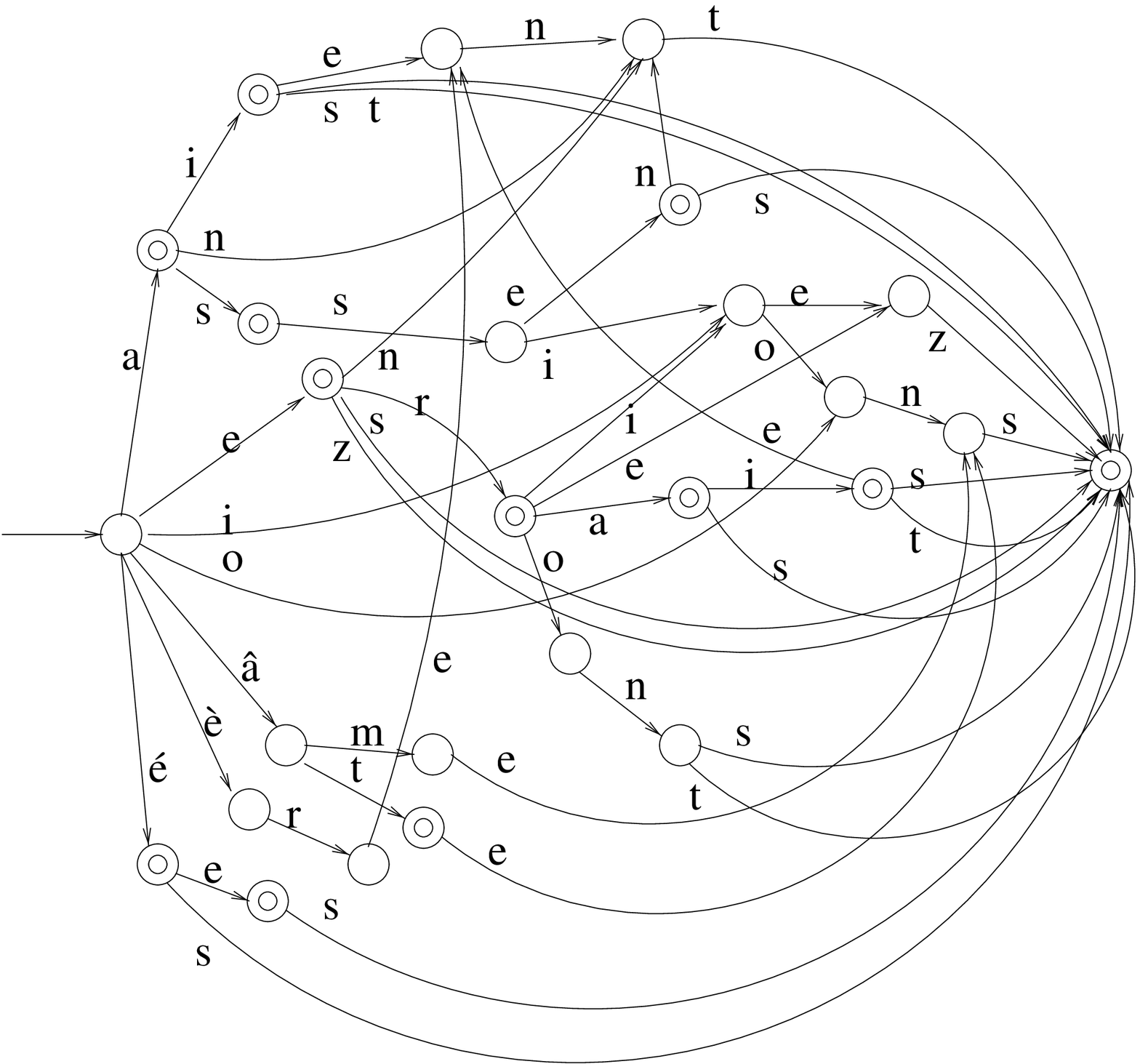,width=0.8\textwidth}}
\caption{The unique minimal dictionary whose language is the French
regular endings of verbs of the first group.}\label{fig:after}
\end{figure}
\end{flushleft}
A trie is a dictionary with a tree-structured transition graph in which the
start state is the root and all leaves are final states\footnote{There may
also be non-leaf, in other words {\em interior}, nodes which are final.}.
An example of a
dictionary in a form of a trie is given in Figure~\ref{fig:before}.
We can see that many subtrees in the transition graph are isomorphic.
The equivalent minimal dictionary (Figure~\ref{fig:after}) is the one
in which only one copy of each isomorphic subtree is kept.
This means that, pointers (edges) to all isomorphic subtrees are replaced by
pointers (edges) to their unique representative.

The traditional method to obtain a minimal dictionary is to first
create a (not necessarily minimal) dictionary for the language and
then minimize it using any one of a number of algorithms (again, see
Watson~\shortcite{Wats93b,Wats95} for numerous examples of such
algorithms).  The first stage is usually done by building a trie, for
which there are fast and well understood algorithms. Dictionary
minimization algorithms are quite efficient in terms of the size of
their input dictionary --- for some algorithms, the memory and time
requirements are both linear in the number of states.  Unfortunately,
even such good performance is not sufficient in practice, where the
intermediate dictionary (the trie) can be much larger than the
available physical memory. (Some effort towards decreasing the memory
requirement has been made; see~\namecite{Revu91}.) This paper presents
a way to reduce these intermediate memory requirements and decrease
the total construction time by constructing the minimal dictionary
incrementally (word by word, maintaining an invariant of minimality),
thus avoiding ever having the entire trie in memory.

The central part of most automata minimization algorithms is a
classification of states.  The states of the input dictionary are
partitioned such that the equivalence classes correspond to the states
of the equivalent minimal automaton.  Assuming the input dictionary
has only reachable
states (that is, $\rpred{Reachable}$ is \nametrue), we can deduce (by
our alternative definition of minimality) that each state in the
minimal dictionary must have a unique right language. Since this is a
necessary and sufficient condition for minimality, we can use equality
of right languages as the equivalence relation for our classes.  Using
our definition of right languages, it is easily shown that equality of
right languages is an equivalence relation (it is reflexive, symmetric
and transitive). We will denote two states, $p$ and $q$, belonging to
the same equivalence class by $p \equiv q$ (note that $\equiv$ here is
different from its use for logical equivalence of predicates). In the
literature, this relation is sometimes written as $E$.


To aid in understanding, let us traverse the trie (see
Figure~\ref{fig:before}) with the postorder method and see how the
partitioning can be performed.
For each state we encounter, we must check whether there is an
equivalent state in the part of the dictionary that has already been
analyzed. If so, we replace the current state with the equivalent
state. If not, we put the state into a register, so that we can find
it easily. It follows that the register has the following property: it
contains only states which are pairwise inequivalent.
We start with the (lexicographically) first leaf, moving backward
through the trie toward the start state. All states up to the first
{\bf forward-branching} state (state with more than one outgoing
transition) must belong to different classes and we immediately place
them in the register, since there will be no need to replace them by
other states. Considering the other branches, and starting from their
leaves, we need to know whether or not a given state belongs to the
same class as a previously registered state.
For a given state $p$ (not in the register), we try to find a state
$q$ in the register that would have the same right language. To do
this, we do not need to compare the languages themselves --- comparing
sets of strings is computationally expensive. We can use our recursive
definition of the right language.
State $p$ belongs to the same class as $q$ if and only if:
\begin{enumerate}
\item \hspace{0.5cm} they are either both final or both non-final; and
\item \hspace{0.5cm} they have the same number of outgoing transitions; and
\item \hspace{0.5cm} corresponding outgoing transitions have the same labels; and
\item \hspace{0.5cm} corresponding outgoing transitions lead to states
  that have the same right \hspace*{0.5cm}languages.
\end{enumerate}
Because the postorder method ensures that all states reachable from
the states already visited are unique representatives of their classes
(i.e.\ their right languages are unique in the visited part of the
automaton), we can rewrite the last condition as:

\begin{enumerate}
\item[4'] \hspace{0.5cm} corresponding transitions lead to the same states.
\end{enumerate}
If all the conditions are satisfied, the state $p$ is
replaced by $q$. Replacing $p$ simply involves deleting it while
redirecting all of its incoming transitions to $q$. Note
that all leaf states belong to the same equivalence class.
If some of the conditions are not satisfied, $p$ must be a
representative of a new class and therefore must be put into the register.

To build the dictionary one word at a time, we need to
merge the process of adding new words to the dictionary with the
minimization process. There are two crucial questions
that need to be answered. First, which states (or equivalence
classes) are subject to change when new words
are added? Second, is there a way to add new words to the dictionary
such that we minimize the number of states that may need to be changed
during the addition of a word? Looking at Figures~\ref{fig:before}
and~\ref{fig:after},
we can reproduce the same postorder traversal of states when the input
data is lexicographically sorted. (Note that in order to do this, the
alphabet $\Sigma$ must be ordered, as is the case with ASCII and
Unicode).  To process a state, we need to know its right language.
According to the method presented above, we must have the whole
subtree whose root is that state. The subtree represents endings of
subsequent (ordered) words. Further investigation reveals that when we
add words in this order, only the states that need to be traversed to
accept the previous word added to the dictionary may change when a new
word is added.  The rest of
the dictionary remains unchanged, because a new word either
\begin{itemize}
\item begins with a symbol different from the first symbols of all
words already in the automaton; the beginning symbol of the new word is
lexicographically placed after those symbols; or
\item it shares some (or even all) initial symbols of the
word previously added to the dictionary; the algorithm then creates a forward
branch, as the symbol on the label of the transition must be
later in the alphabet than symbols on all other transitions leaving
that state.
\end{itemize}
When the previous word is a prefix of the new
word, the only state that is to be modified is the last state
belonging to the previous word. The new word may share
its ending
with other words already in the dictionary, which
means that we need to create links to some parts of the dictionary.
Those parts, however, are not modified. This discovery leads us to the
Algorithm~\ref{algorithm-sorted}, shown below.

\begin{tabbing}
  \hspace{0.6cm}\=\kill \\
  {\bf Algorithm \algcounter\label{algorithm-sorted}} \\
  
  \hrulefill
  \ \\
  \> {\it Register\/} := $\emptyset$; \\
  \> {\bf do} \=\mbox{there is another word} $\rightarrow$ \\
  \>\> {\it Word\/} := next word in lexicographic order; \\
  \>\> {\it CommonPrefix} := {\it common\_prefix(Word)\/}; \\
  \>\> {\it LastState\/} := $\delta^*(q_0,{\mathit CommonPrefix})$; \\
  \>\> {\it CurrentSuffix} := {\it Word[length(CommonPrefix)+1\ldots{}length(Word)]}; \\
  \>\> {\bf if} \={\it has\_children(LastState)} $\rightarrow$ \\
  \>\>\> {\it replace\_or\_register(LastState)} \\
  \>\> {\bf fi}; \\
  \>\> {\it add\_suffix(LastState, CurrentSuffix)} \\
  \> {\bf od}; \\
  \> {\it replace\_or\_register($q_0$)} \\
  \ \\
  \> {\bf func} \={\it common\_prefix(Word)} $\rightarrow$ \\
  \>\> {\bf return} the longest prefix $w$ of {\it Word\/} such that $\delta^*(q_0,w) \neq \bot$ \\
  \> {\bf cnuf} \\
  \ \\
  \> {\bf func} \={\it replace\_or\_register(State)} $\rightarrow$ \\
  \>\> {\it Child} := {\it last\_child(State)\/}; \\
  \>\> {\bf if} \={\it has\_children(Child)} $\rightarrow$ \\
  \>\>\> {\it replace\_or\_register(Child)} \\
  \>\> {\bf fi}; \\
  \>\> {\bf if} \=$\exists_{q\in{}Q}(q \in \mbox{\it Register} \wedge q
  \equiv \mbox{\it Child}) \rightarrow$ \\
  \>\>\> {\it last\_child(State)} := $q : (q \in \mbox{\it Register} \land q
        \equiv \mbox{\it Child})$; \\
  \>\>\> {\it delete(Child)} \\
  \>\> {\bf else} \\
  \>\>\> {\it Register} := $\mbox{\it Register} \cup \{\mbox{\it Child}\}$ \\
  \>\> {\bf fi} \\
  \> {\bf cnuf} \\
  \ \\
\end{tabbing}

The main loop of the algorithm reads subsequent words and establishes
which part of the word is already in the automaton (the {\it
  CommonPrefix\/}), and which is not (the {\it CurrentSuffix\/}). An
important step is determining what the last state (here called {\it
  LastState\/}) in the path of the common prefix is. If {\it LastState\/}
already has children, it means that not all states in the path of
previously added word are in the path of the common prefix. In that case, by
calling the function \rmathfunc{replace\_or\_register}, we can
let the minimization process work on those states in the path of the
previously added word that are not in the common prefix path. Then we
can add to the {\it LastState\/} a chain of states that would recognize
the {\it CurrentSuffix}.

The function \rmathfunc{common\_prefix} finds the longest prefix (of the word
to be added) that is a prefix of a word already in the automaton.
The prefix can be empty (since $\delta^*(q, \emptystring)=q$).

The function \rmathfunc{add\_suffix} creates a branch extending out of the
dictionary, which represents the suffix of the word being added (the
maximal suffix of the word which is not a prefix of any other word
already in the dictionary). The last state of this branch is marked as
final.

The function \rmathfunc{last\_child} returns a
reference to the state reached by the
lexicographically last transition that is outgoing from the argument
state. Since the input data is lexicographically sorted, {\em
last\_child\/} returns the outgoing transition (from the state)
most recently added (during the addition of the previous word).
The function \rmathfunc{replace\_or\_register}
effectively works on the last child of the argument state. It is
called with the argument that is the last state in the common prefix
path (or the initial state in the last call). We need the argument
state to modify its transition in those instances in which the child is to
be replaced with another (equivalent) state. Firstly, the function
calls itself recursively until it reaches the end of the path of the
previously added word. Note that when it encounters a state with more
than one child, it takes the last one, as it belongs to the previously
added word. As the length of words is limited, so is the depth of
recursion.  Then, returning from each recursive call, it checks
whether a state equivalent to the current state can be found in the
register. If this is true, then the state is replaced with the
equivalent state found in the register. If not, the state is
registered as a representative of a new class. Note that the function
\rmathfunc{replace\_or\_register} processes only the states belonging
to the path of the previously added word (a part, or possibly all of
those created with the previous call to \rmathfunc{add\_suffix}), and
that those states are never reprocessed.
Finally, \rmathfunc{has\_children} returns
\nametrue\ if, and only if, there are outgoing transitions from the
state.

During the construction, the automaton states are either in the
register or on the path for the last added word. All the states in the
register are states in the resulting minimal automaton. Hence the
temporary automaton built during the construction has less states than
the resulting automaton plus the
length of the longest word.
Memory is needed for the minimized dictionary that is
under construction, the call stack, and for the register of states. The
memory for the dictionary is proportional to the number of states and
the total number of transitions. The memory for the register of states
is proportional to the number of states and can be freed once
construction is complete.
By choosing an appropriate implementation method one can achieve a memory 
complexity $\bigoh{n}$ for a given alphabet, where $n$ is the number
of states of the minimized automaton. This is an important advantage
of our algorithm.

For each letter from the input list, the algorithm either has to make
a step in the function \rmathfunc{common\_prefix} or add a state in
the procedure \rmathfunc{add\_sufix}. Both operations can be performed
in constant time. Each new state that has been added in the procedure
\rmathfunc{add\_sufix} has to be processed exactly once in the
procedure \rmathfunc{replace\_or\_register}. The number of states that
have to be replaced or registered is clearly smaller than the number
of letters in the input list\footnote{The exact number of the states
  that are processed in the procedure
  \rmathfunc{replace\_or\_register} is equal to the number of states
  in the trie for the input language.}. The processing of one state in
the procedure consists of one register search and possibly one
register insertion.  The time complexity of the search is $\bigoh{\log
  n}$,where $n$ is the number of states in the (minimized) dictionary.
The time complexity of adding a state to the register is also
$\bigoh{\log n}$. In practice, however, by using a hash table to
represent the register (and equivalence relation), the average time
complexity of those operations can be made almost constant. Hence the
time complexity of the whole algorithm is $\bigoh{l \log n}$, where
$l$ is the total number of letters in the input list.


\section{Construction from Unsorted Data}
Sometimes it is difficult or even impossible to sort the input data
before constructing a dictionary. For example, there may be
insufficient time or storage space to sort the data or the data
originates in another program or physical source. An incremental
dictionary-building algorithm would still be very useful in those
situations, although unsorted data makes it more difficult to merge
the trie-building and the minimization processes. We could leave the
two processes disjoint, although this would lead to the traditional
method of constructing a trie and minimizing it afterwards. A better
solution is to minimize everything on-the-fly, possibly changing the
equivalence classes of some states each time a word is added.  Before
actually constructing a new state in the dictionary, we first
determine if it would be included in the equivalence class of a
preexisting state. Similarly, we may need to change the equivalence
classes of previously constructed states since their right languages
may have changed. This leads to an incremental construction
algorithm. Naturally, we would want to create the states for a new
word in an order that would minimize the creation of new equivalence
classes.

\begin{figure}[htb]
\begin{center}
\begin{minipage}[b]{0.45\textwidth}
\mbox{\psfig{figure=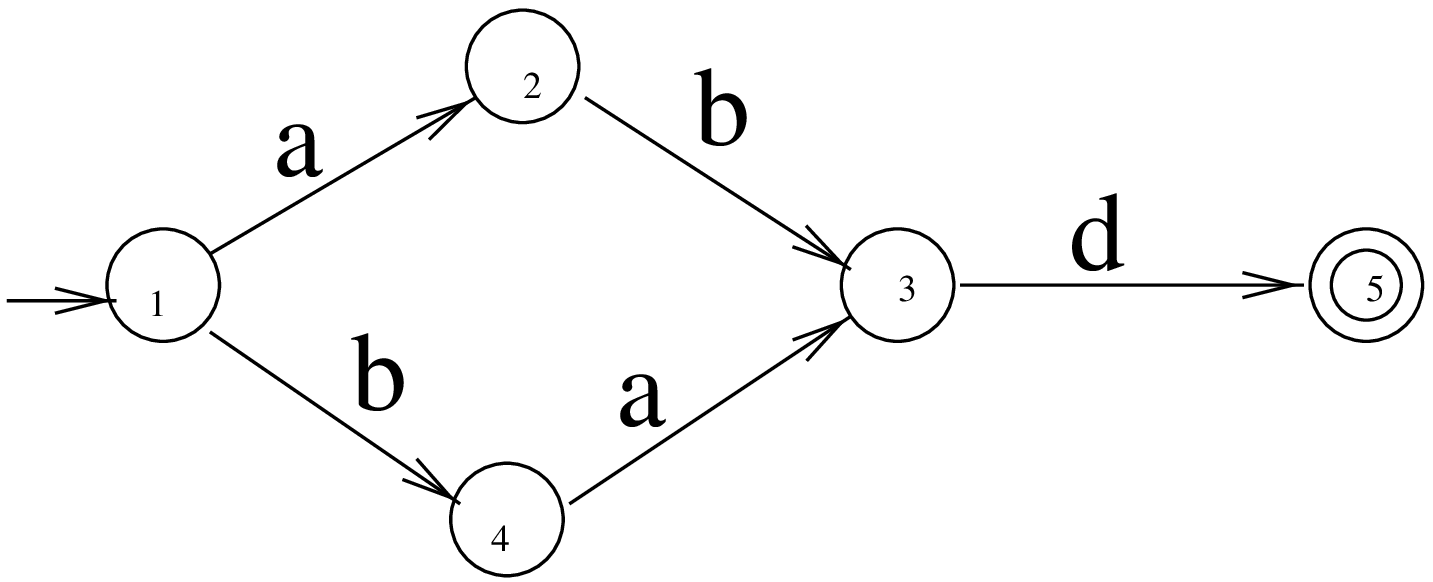,width=0.9\textwidth}}
\end{minipage}\hfill%
\begin{minipage}[b]{0.45\textwidth}
\mbox{\psfig{figure=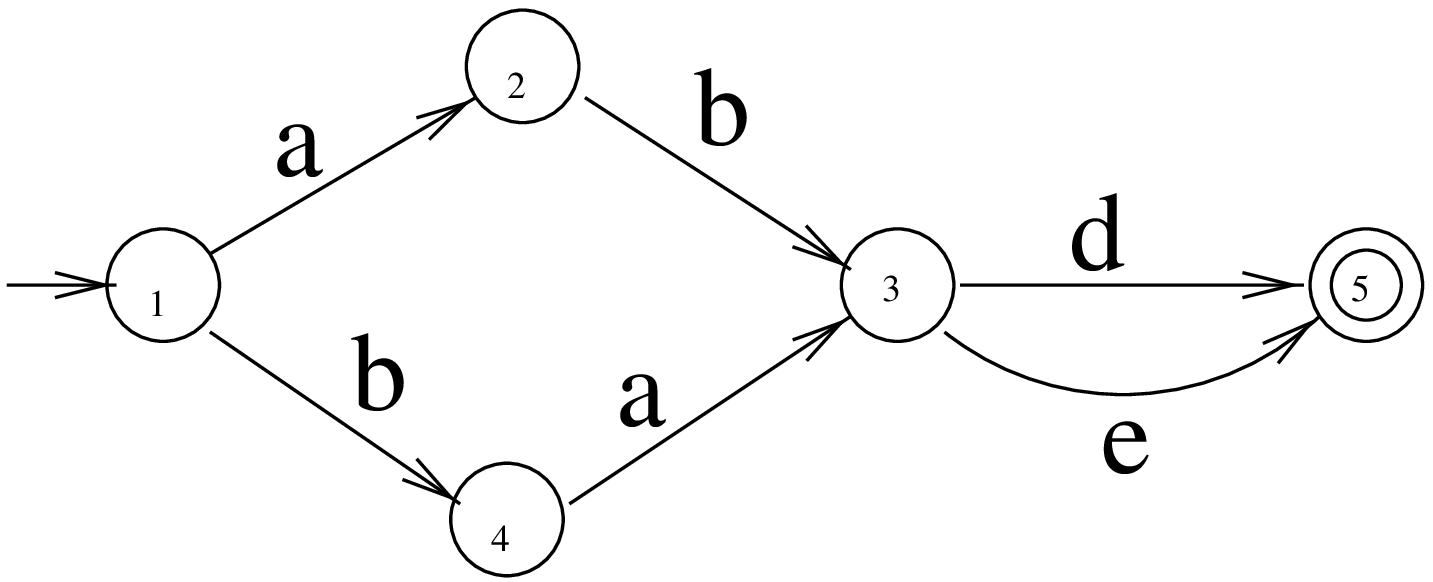,width=0.9\textwidth}}
\end{minipage}
\begin{minipage}[b]{0.45\textwidth}
\mbox{\psfig{figure=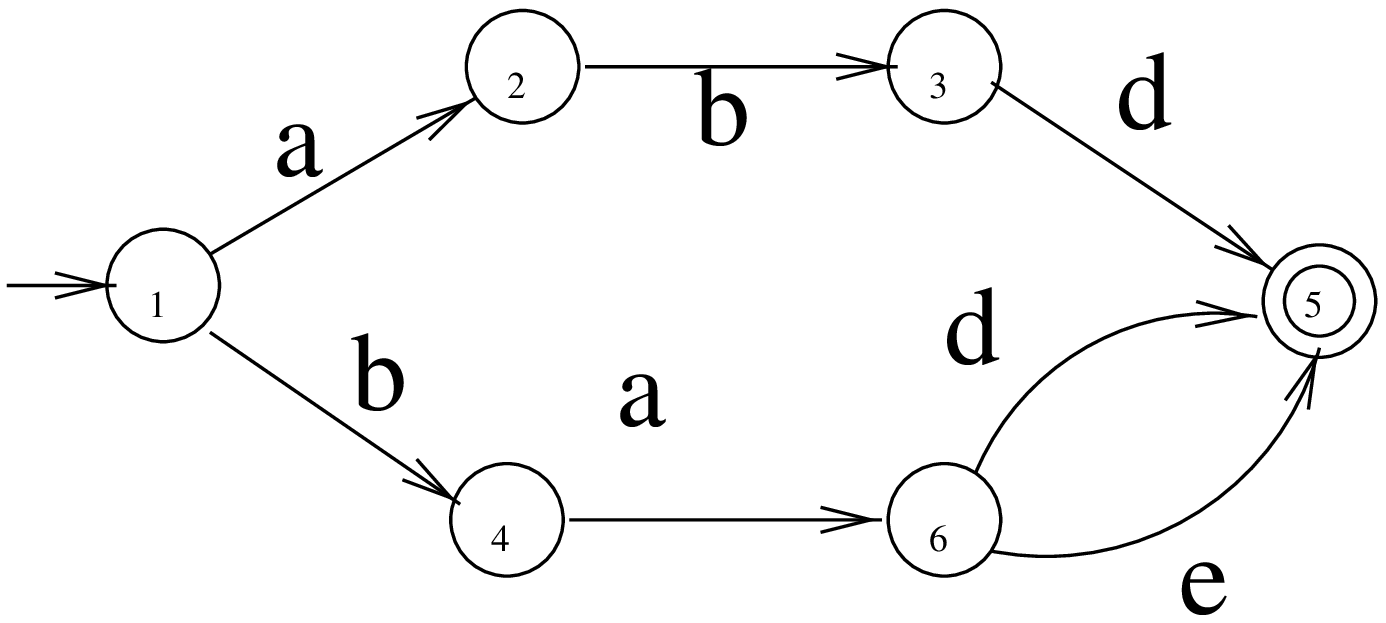,width=0.9\textwidth}}
\end{minipage}
\caption{The result of blindly adding the word {\em bae\/} to a minimized
dictionary (appearing on the left) containing {\em abd\/} and {\em bad}. The
rightmost dictionary inadvertently contains {\em abe\/} as well. The lower
dictionary is correct --- state $3$ had to be cloned.
}\label{fig:multiply}
\end{center}
\end{figure}

As in the algorithm for sorted data, when a new word $w$ is added, we
search for the prefix of $w$ already in the dictionary. This time,
however, we cannot assume that the states traversed by this common
prefix will not be changed by the addition of the word. If there are
any preexisting states traversed by the common prefix that are
already targets of more than one in-transition (known as {\bf
confluence\/} states), then blindly appending another transition to
the last state in this path (as we would in the sorted algorithm)
would accidentally add more words than desired (see
Figure~\ref{fig:multiply} for an example of this).

To avoid generation of such spurious words, all states in the common
prefix path from the first confluence state
must be {\bf cloned}. Cloning is the process of creating a new state
that has outgoing transitions on the same labels and to the same
destination states as a given state.  If we compare the minimal
dictionary (Figure~\ref{fig:before}) to an equivalent trie
(Figure~\ref{fig:after}), we notice that a confluence state can be
seen as a root of several original, isomorphic subtrees merged into
one (as described in the previous section).  One of the isomorphic
subtrees now needs to be modified (leaving it no longer isomorphic),
so it must first be separated from the others by cloning its root.
The isomorphic subtrees hanging off these roots are unchanged, so the
original root and its clone have the same outgoing transitions (that
is, transitions on the same labels and to the same destination
states).

In the algorithm~\ref{algorithm-sorted}, the confluence states were
never traversed during the search for the common prefix. The common
prefix was not only the longest common prefix of the word to be added
and all the words already in the automaton. It was also the longest
common prefix of the word to be added and the last (i.e.\ the
previous) word added to the automaton. As it was the function
\rmathfunc{replace\_or\_register} that created confluence states, and
that function was never called on states belonging to the path of the
last word added to the automaton, those states could never be found in
the common prefix path.

Once the entire common prefix is traversed, the rest of the word must
be appended. If there are no confluence states in the common prefix,
then the method of adding the rest of the word does not differ from
the method used in the algorithm for sorted data.
However, we need to withdraw (from the register) the last state in the
common prefix path in order not to create cycles. This is in contrast
to the situation in the algorithm for sorted data where that state is
not yet registered. Also, {\it CurrentSuffix\/} could be matched with
a path in the automaton containing states from the common prefix path
(including the last state of the prefix).

\begin{figure}[htb]
\begin{center}
\begin{minipage}[b]{0.49\textwidth}
\mbox{\psfig{figure=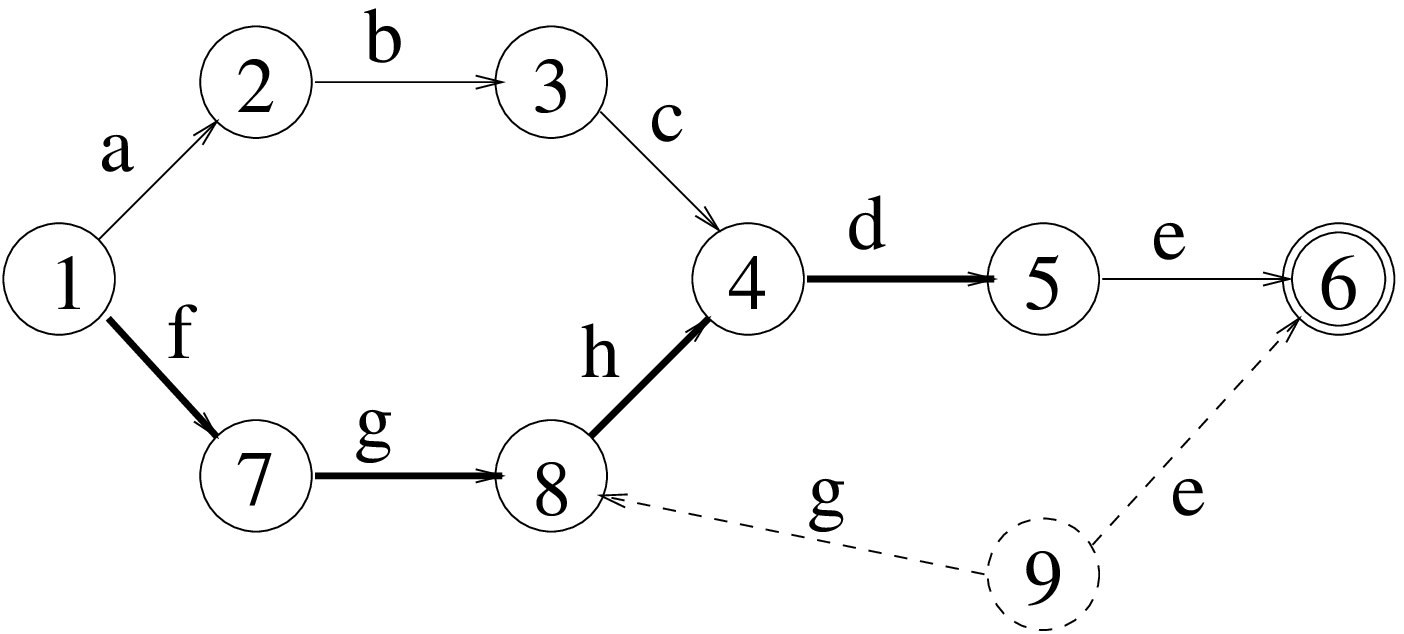,width=0.9\textwidth}}
\end{minipage}\hfill%
\begin{minipage}[b]{0.49\textwidth}
\mbox{\psfig{figure=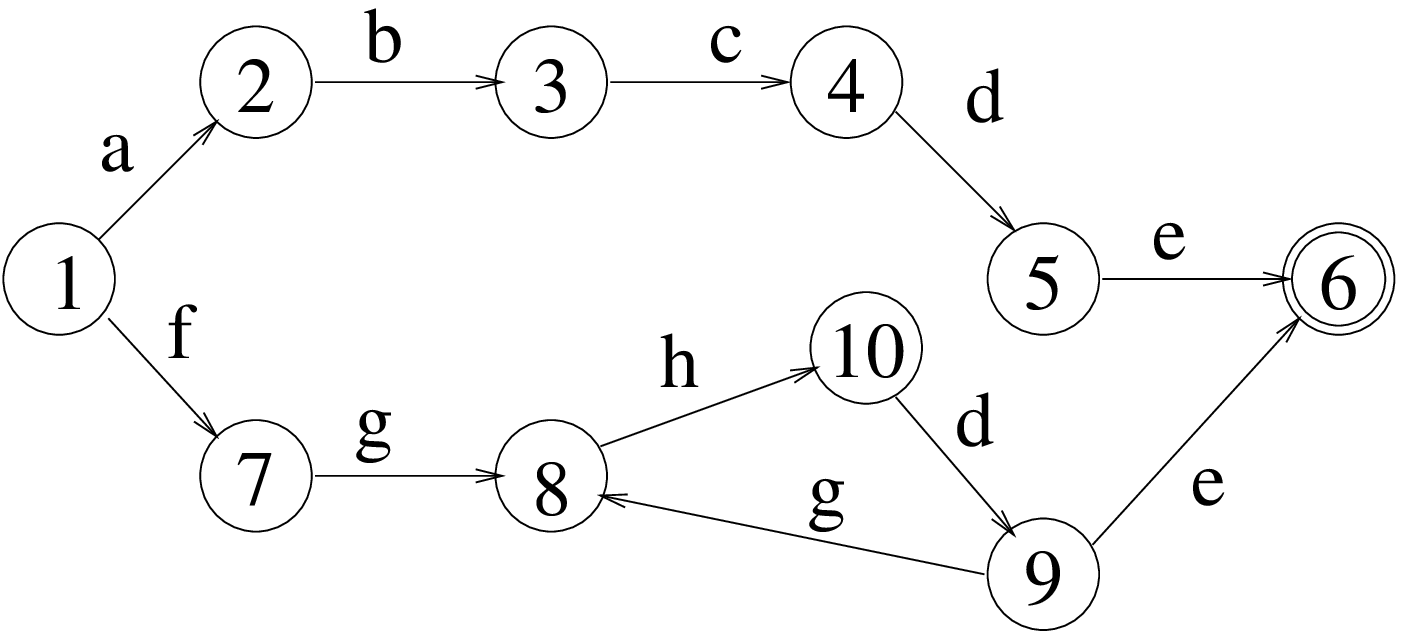,width=0.9\textwidth}}
\end{minipage}
\caption{Consider an automaton (shown in solid lines on the left)
  accepting {\em abcde\/} and {\em fghde}. Suppose we want to
  add {\em fghdghde}. As the common prefix path (shown in thicker lines)
  contains a confluence state, we clone state 5 to obtain state 9,
  add the suffix to state 9, and minimize it. When we also consider the dashed
  lines
  in left figure, we see that state 8 became a new confluence state earlier in
  the common prefix path. The right figure shows what could happen if
  we did not rescan the common prefix path for confluence
  states. State 10 is a clone of state 4.}\label{fig:prepost}
\end{center}
\end{figure}

When there is a confluence state, then we need to clone some
states. We start with the last state in the common prefix path,
append the rest of the word to that clone and minimize it. Note that
in this algorithm, we do not wait for the next word to come, so we can
minimize (replace or register the states of) {\it CurrentSuffix\/} state
by state as they are created. Adding and minimizing the rest of the
word may create new confluence states earlier in the common prefix
path, so we need to rescan the common prefix path in order not to
create cycles, as illustrated in Figure~\ref{fig:prepost}.
Then we proceed with cloning and minimizing the states on the path from the
state immediately preceding the last state to the current first confluence
state.

Another, less complicated but also less economical, method can be used
to avoid the problem of creating cycles in presence of confluence
states. In that solution, we proceed from the state immediately
preceding the confluence state towards the end of the common prefix
path, cloning the states on the way. But first, the state
immediately preceding the first confluence state should be removed
from the register. At the end of the common prefix path, we add the
suffix. Then, we call \rmathfunc{replace\_or\_register} with the
predecessor of the state immediately preceding the first confluence
state. The following should be noted about this solution:
\begin{itemize}
\item memory requirements are higher, as we keep more than one isomorphic state
at a time,
\item the function \rmathfunc{replace\_or\_register} must remain
recursive (as in the sorted version), and
\item the argument to \rmathfunc{replace\_or\_register} must be a
string, not a symbol, in order to pass subsequent symbols to children.
\end{itemize}

When the process of traversing the common prefix (up to a confluence
state) and adding the suffix is complete, further modifications follow.
We must recalculate the equivalence class of each state on the
path of the new word. If any equivalence class changes, we must also
recalculate the equivalence classes of all of the parents of all of the
states in the changed class.
Interestingly, this process could actually make the new dictionary smaller.
For example, if we add the word {\em abe\/} to the dictionary at the bottom of
Figure~\ref{fig:multiply} while maintaining minimality, we
obtain the dictionary shown in the right of Figure~\ref{fig:multiply}, which
is one state smaller.
The resulting algorithm is shown in Algorithm~\ref{algorithm-unsorted}.
\begin{tabbing}
  \hspace{0.6cm}\=\kill \\
  {\bf Algorithm \algcounter\label{algorithm-unsorted}} \\
  \ \\
  \> {\it Register\/} := $\emptyset$; \\
  \> {\bf do} \=\mbox{there is another word} $\rightarrow$ \\
  \>\> {\it Word\/} := \mbox{next word}; \\
  \>\> {\it CommonPrefix} := {\it common\_prefix}$(\mbox{\it Word})$; \\
  \>\> {\it CurrentSuffix} :=
      {\it Word[length(CommonPrefix)+1\ldots{}length(Word)]}; \\
  \>\> {\bf if} \=${\it CurrentSuffix} = \emptystring \land \delta^*(q_0,
        {\it CommonPrefix}) \in F \rightarrow$ \\
  \>\>\> {\bf continue} \\
  \>\> {\bf fi}; \\
  \>\> {\it FirstState} := {\it first\_state}$(\mbox{\it CommonPrefix})$; \\
  \>\> {\bf if} \={\it FirstState} $= \emptyset \rightarrow$ \\
  \>\>\> {\it LastState\/} := $\delta^*(q_0, \mbox{\it CommonPrefix})$ \\
  \>\> {\bf else} \\
  \>\>\> {\it LastState} := $\mbox{\it clone}(\delta^*(q_0,\mbox{\it CommonPrefix}))$ \\
  \>\> {\bf fi}; \\
  \>\> {\it add\_suffix\/}({\it LastState, CurrentSuffix\/}); \\
  \>\> {\bf if} \={\it FirstState} $\not=\emptyset \rightarrow$ \\
  \>\>\> {\it FirstState} := {\it first\_state\/}({\it CommonPrefix\/}); \\
  \>\>\> {\it CurrentIndex} := {\it (length(x)}$: \delta^*(q_0,x) =$ {\it
        FirstState\/}); \\
  \>\>\> {\bf for} \=$i$ {\bf from} {\it length(CommonPrefix) - 1} {\bf
  downto} {\it CurrentIndex} $\rightarrow$ \\
  \>\>\>\> {\it CurrentState} := $\mbox{\it clone}(\delta^*(q_0,
    \mbox{\it CommonPrefix[1\ldots{}i]}))$; \\
  \>\>\>\> $\delta(\mbox{\it CurrentState, CommonPrefix[i]})$ := {\it
  LastState\/}; \\
  \>\>\>\> {\it replace\_or\_register\/}({\it CurrentState},{\it Word[i+1]\/}); \\
  \>\>\>\> {\it LastState} := {\it CurrentState} \\
  \>\>\> {\bf rof} \\
  \>\> {\bf else} \\
  \>\>\> {\it CurrentIndex} := {\it length\/}({\it CommonPrefix}) \\
  \>\> {\bf fi}; \\
  \>\> {\it Changed} := {\it true\/}; \\
  \>\> {\bf do} \={\it Changed} $\rightarrow$ \\
  \>\>\> {\it CurrentIndex} := {\it CurrentIndex - 1\/}; \\
  \>\>\> {\it CurrentState\/} :=
    $\delta^*(q_0, \mbox{\it Word[1\ldots{}CurrentIndex]})$; \\
  \>\>\> {\it OldState} := {\it LastState\/}; \\
  \>\>\> {\bf if} $\mbox{\it CurrentIndex} > 0 \rightarrow$ \\
  \>\>\>\> {\it Register} := {\it Register} - $\{{\mathit LastState}\}$ \\
  \>\>\> {\bf fi}; \\
  \>\>\> {\it replace\_or\_register\/}({\it CurrentState},{\it
  Word[CurrentIndex+1]\/}); \\
  \>\>\> {\it LastState} := $\delta$({\it CurrentState}, {\it
  Word[CurrentIndex+1]\/});\\
  \>\>\> {\it Changed} := {\it OldState} $\not=$ {\it LastState} \\
  \>\>\> {\it LastState\/} := {\it CurrentState\/} \\
  \>\> {\bf od} \\
  \>\> {\bf if} $\lnot \mbox{\it Changed\/} \land \mbox {\it CurrentIndex\/}
        > 0 \rightarrow$ \\
  \>\>\> {\it Register} := {\it Register\/} $\cup$ $\{\mbox{\it CurrentState}\}$ \\
  \>\> {\bf fi} \\
  \> {\bf od} \\
  \ \\
  \> {\bf func} \={\it replace\_or\_register\/}({\it State, Symbol}) $\rightarrow$ \\
  \>\> {\it Child\/} := $\delta(${\it State, Symbol}$)$; \\
  \>\> {\bf if} \=$\exists{q\in{}Q}(q \in \mbox{\it Register} \land q
  \equiv {\mathit Child}) \rightarrow$ \\
  \>\>\> {\it delete\/}({\it Child\/}); \\
  \>\>\> {\it last\_child\/}({\it State\/}) := $q : (q \in \mbox{\it
    Register} \land q \equiv \mbox{\it Child})$ \\
  \>\> {\bf else} \\
  \>\>\> {\it Register} := {\it Register} $\cup \{${\it Child\/}$\}$ \\
  \>\> {\bf fi} \\
  \> {\bf cnuf} \\
\end{tabbing}

The main loop reads the words, finds the common prefix, and tries to
find the first confluence state in the common prefix path. Then the
remaining part of the word ({\it CurrentSuffix\/}) is added.

If a confluence state is found (i.e.\ {\it FirstState\/} points to a
state in the automaton), all states from the first confluence state to
the end of the common prefix path are cloned, and then considered for
replacement or registering. Note that the inner loop (with {\it i\/}
as the control variable) begins with the penultimate state in the
common prefix, because the last state has already been cloned and the
function \rmathfunc{replace\_or\_register} acts on a child of its
argument state.

Addition of a new suffix to the last state in the common prefix
changes the right languages of all states that precede that state in
the common prefix path. The last part of the main loop deals with that
situation. If the change resulted in such modification of the right
language of a state that an equivalent state can be found somewhere
else in the automaton, then the state is replaced with the equivalent
one and the change propagates towards the initial state. If the
replacement of a given state cannot take place, then (according to our
recursive definition of the right language) there is no need to
replace any preceding state.

Several changes to the functions used in the sorted algorithm are
necessary to handle the general case of unsorted data. The
\rmathfunc{replace\_or\_register} procedure
needs to be modified slightly. Since new words are added in arbitrary
order, one can no longer assume that the last child (lexicographically)
of the state (the one that has been added most recently) is the
child whose equivalence class may have changed.
However, we know the label on the transition leading to the altered
child, so we use it to access that state. Also, we do not need to call
the function recursively.  We assume that \rmathfunc{add\_suffix}
replaces or registers the states in the {\it CurrentSuffix\/} in
correct order; later we process one path of states in the automaton,
starting from those most distant from the initial state, proceeding
towards the initial state $q_0$.  So in every situation in which we
call \rmathfunc{replace\_or\_register}, all children of the state {\it
  Child\/} are already unique representatives of their equivalence
classes.

Also, in the sorted algorithm, \rmathfunc{add\_suffix} is never passed
$\emptystring$ as an argument, whereas this may occur in the unsorted
version of the algorithm. The effect is that the \rvar{LastState}
should be marked as final since the common prefix is, in fact, the
entire word.
In the sorted algorithm, the chain of states created by
\rmathfunc{add\_suffix} was left for further treatment until new words
are added (or until the end of processing). Here, the automaton is
completely minimized on the fly after adding a new word, and the
function \rmathfunc{add\_suffix} can call
\rmathfunc{replace\_or\_register} for each state it creates (starting
from the end of the suffix).
Finally, the new function \rmathfunc{first\_state} simply traverses the
dictionary using the given word prefix and returns the first confluence state
it encounters.
If no such state exists, \rmathfunc{first\_state} returns $\emptyset$.

As in the sorted case, the main loop of the unsorted algorithm
executes $m$ times, where $m$ is the number of words accepted by the
dictionary. The inner loops are executed at most $|w|$ times for each
word. Putting a state into the register takes $\bigoh{\log n}$,
although it may be constant when using a hash table. The same
estimation is valid for a removal from the register. In this case, the
time complexity of the algorithm remains the same, but the constant
changes. Similarly, hashing can be used to provide an efficient method
of determining the state equivalence classes. For sorted data, only a
single path through the dictionary could possibly be changed each time
a new word is added. For unsorted data, however, the changes
frequently fan-out and percolate all the way back to the start state,
so processing each word takes more time.

\subsection{Extending the algorithms}
These new algorithms can also be used to construct transducers. The
alphabet of the (transducing) automaton would be $\Sigma_1 \times
\Sigma_2$, where $\Sigma_1$ and $\Sigma_2$ are the alphabet of the
levels. Alternatively, elements of $\Sigma_{2}^{*}$ can be associated
with the final states of the dictionary and only output once a valid
word from $\Sigma_{1}^{*}$ is recognized.

\section{Related work}
An algorithm described by \namecite{Revu91} also constructs a dictionary
from sorted data while performing a partial minimization on-the-fly.
Data is sorted in reverse order
and that property is used to compress the endings of words within the
dictionary as it is being built.  This is called a {\bf
  pseudominimization} and must be supplemented by a true minimization
phase afterwards. The minimization phase still involves finding an
equivalence relation over all of the states of the pseudo-minimal
dictionary. It is possible to use unsorted data but it produces a much
bigger dictionary in the first stage of processing.  However, the time
complexity of the minimization can be reduced somewhat by using
knowledge of the pseudo-minimization process.  Although this
pseudo-minimization technique is more economic in its use of memory
than traditional techniques, we are still left with a sub-minimal
dictionary which can be a factor of 8 times larger than the equivalent
minimal dictionary (\cite[page 33]{Revu91}, reporting on the DELAF
dictionary).

Recently, a semi-incremental algorithm was described by
\namecite{Wats98} at the Workshop on Implementing Automata. That
algorithm requires the words to be sorted in {\em any\/} order of
decreasing length (this sorting process can be done in linear time),
and takes advantage of similar automata properties to those presented
in this paper. In addition, the algorithm requires a {\em final\/}
minimization phase after all words have been added. For this reason,
it is only semi-incremental and does not maintain full minimality
while adding words --- although it usually maintains the automata
close enough to minimality for practical applications.

\section{Conclusions}
We have presented two new methods for incrementally constructing a minimal,
deterministic, acyclic finite state automaton from a finite set of words
(possibly with corresponding annotations).
Their main advantage is their minimal intermediate memory
requirements\footnote{It is minimal in asymptotic terms; naturally compact data
structures can also be used.}.
The total construction time of these
minimal dictionaries is dramatically reduced from previous algorithms.
The algorithm constructing a dictionary from sorted data can be used
in parallel with other algorithms that traverse or utilize the
dictionary since parts of the dictionary that are already constructed are
no longer subject to future change.

\newpage

\starttwocolumn

\sloppy

\section{Acknowledgements}
Jan Daciuk would like to express his gratitude to the Swiss Federal
Scholarship Commission for providing a scholarship that made possible
the work described here. Jan would also like to thank friends from ISSCO,
Geneva, for their comments and suggestions on early versions of the algorithms
given in this paper.

Bruce Watson and Richard Watson would like to
thank Ribbit Software Systems Inc.\ for its continued support in
these fields of applicable research.

All authors would like to thank the anonymous reviewers and Nanette Saes for
their valuable comments and suggestions that led to significant improvements
of the paper.

\bibliographystyle{fullname}

\bibliography{daciuk98}

\end{document}